\documentclass[sigconf]{acmart}

\usepackage{dblfloatfix}
\usepackage{multirow}

\def\ie{{\em i.e.}}
\def\eg{{\em e.g.}}

\def\BibTeX{{\rm B\kern-.05em{\sc i\kern-.025em b}\kern-.08emT\kern-.1667em\lower.7ex\hbox{E}\kern-.125emX}}
    
\copyrightyear{2019}
\acmYear{2019}
\setcopyright{acmlicensed}
\acmConference[]{}{}{}
\acmBooktitle{}
\acmPrice{}
\acmDOI{}
\acmISBN{}

\begin{document}

%
% The "title" command has an optional parameter, allowing the author to define a "short title" to be used in page headers.
\title{Exposing GAN-synthesized Faces Using Landmark Locations}

%
% The "author" command and its associated commands are used to define the authors and their affiliations.
% Of note is the shared affiliation of the first two authors, and the "authornote" and "authornotemark" commands
% used to denote shared contribution to the research.
\author{Xin Yang$^\ast$, Yuezun Li$^\ast$, Honggang Qi$^\dagger$, Siwei Lyu$^\ast$}

\affiliation{%
  \institution{	$^\ast$ Computer Science Department, University at Albany, State University of New York, USA}
    \institution{$^\dagger$ School of Computer and Control Engineering, University of the Chinese Academy of Sciences, China}
}

%
% By default, the full list of authors will be used in the page headers. Often, this list is too long, and will overlap
% other information printed in the page headers. This command allows the author to define a more concise list
% of authors' names for this purpose.
\renewcommand{\shortauthors}{}

%
% The abstract is a short summary of the work to be presented in the article.
\begin{abstract}
Generative adversary networks (GANs) have recently led to highly realistic image synthesis results. In this work, we describe a new method to expose GAN-synthesized images using the locations of the facial landmark points. Our method is based on the observations that the facial parts configuration generated by GAN models are different from those of the real faces, due to the lack of global constraints. We perform experiments demonstrating this phenomenon, and show that an SVM classifier trained using the locations of facial landmark points is sufficient to achieve good classification performance for GAN-synthesized faces.

\end{abstract}


%
% A "teaser" image appears between the author and affiliation information and the body 
% of the document, and typically spans the page. 
%   \Description{Enjoying the baseball game from the third-base seats. Ichiro Suzuki preparing to bat.}

%
% This command processes the author and affiliation and title information and builds
% the first part of the formatted document.
\maketitle

\section{Introduction}
\label{sec:intro}
\begin{figure*}[t]
	\centering
	\includegraphics[width=0.8\linewidth]{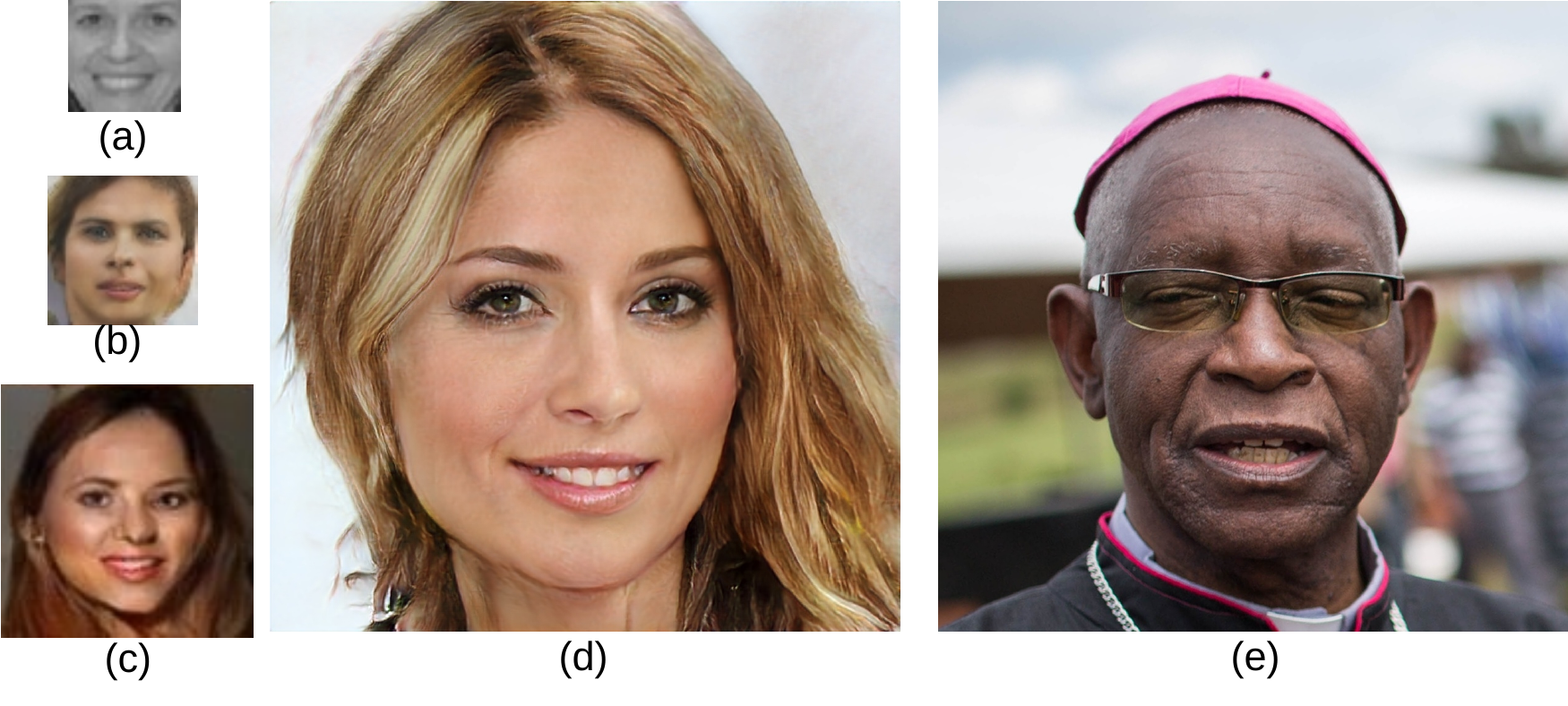}
	~\vspace{-1.5em}
	\caption{\small \em {Over the years, GAN models have been improved significantly over the quality of faces they synthesize. Here we show a few examples of different GAN models (a) GAN \cite{goodfellow2014generative}, (b) DCGAN \cite{radford2015unsupervised}, (c) COGAN \cite{liu2016coupled}, (d) PGGAN \cite{karras2017progressive}, (e) Style-GAN \cite{karras2018style}.}}
	~\vspace{-2em}
	\label{fig:gan_faces}
	~\vspace{1.5em}
\end{figure*}

The fast advancement of artificial intelligence technologies and the increasing availability of large volume of online images and videos and high-throughput computing hardware have revolutionized the tools to generate visually realistic images and videos. These technologies are becoming more efficient and accessible to more users. The recent developments in deep neural networks \cite{Goodfellow-et-al-2016, deng2014tutorial, zhang2018survey, arulkumaran2017brief}, and in particular, the generative adversary networks (GANs) \cite{goodfellow2014generative}, have spawned a new type of image synthesis methods that can produce images with high levels of realism. Figure \ref{fig:gan_faces} shows a few examples of GAN synthesized faces, with very impressive results obtained using recent GAN-based methods \cite{karras2017progressive, karras2018style}.  

The increasing sophistication of GAN-synthesized images also has the negative effect of  fake visual media, and the most damaging examples of which are perhaps the fabricated or manipulated human faces since faces carry the most identifiable information of a person.  The wide spread of fake media with GAN-synthesized faces raise significant ethical, legal and security concerns, and there is an urgent need for methods that can detect GAN-synthesized faces in images and videos.

Unlike previous image/video manipulation methods, realistic images are generated completely from random noise through a deep neural network. Current detection methods are based on low level features such as color disparities \cite{li2018detection, mccloskey2018detecting}, or using the whole image as input to a neural network to extract holistic features \cite{tariq2018detecting}. 

In this work, we develop a new GAN-synthesized face detection method based on a more semantically meaningful features, namely the locations of facial landmark points. This is because the GAN-synthesized faces exhibit certain abnormality in the facial landmark locations. Specifically, The GAN-based face synthesis algorithm can generate face parts (\eg, eyes, nose, skin, and mouth, etc) with a great level of realistic details, yet it does not have an explicit constraint over the locations of these parts in a face. To make an analogy, the GAN-based face synthesis method works like players in a game of Fukuwarai\footnote{Fukuwarai is a traditional game played in Japan during the new year time. A player of Fukuwarai is blindfolded and is requested to put parts of the face (\ie, the eyes, eyebrows, nose and mouth), usually printed on paper, onto a blank face.}, it has all the face parts, but lacks in placing them in a natural and coherent way as in a real face. 

We show that these abnormalities in the configuration of facial parts in GAN-synthesized faces can be revealed using the locations of the facial landmark points (\eg, tips of the eyes, nose and the mouth) automatically detected on faces. To accommodate the variations in shape, orientation and scale of different faces, we further normalize all the facial landmarks to the same standard coordinate system. We then used the normalized locations of these facial landmarks as features for a simple SVM classifier. The landmark location based SVM classifier is tested on faces generated with the state-of-the-art GAN-based face synthesis PGGAN \cite{karras2017progressive} where it shows reasonable classification performance while only using low dimensional features and a light model with fewer parameters.

%\yx{(article about concerns , cite this article? https://www.pcmag.com/news/366592/this-site-is-freaking-people-out-with-ai-generated-lifelike)}

%which assume the manipulated images or videos are obtained through editing tools and manual processing. For instance, previous image manipulation detections tend to look for artifacts left by image editing software (\eg, Photoshop) or camera characteristics as cues for detection, such as display/image distortion \cite{patel2016secure, boulkenafet2016face}, splicing inconsistency in tampered regions \cite{korus2016multi, li2017image}. Also, recent methods developed for detecting automatically generated ``DeepFake'' videos use inconsistencies between real imagehead poses inconsistency \cite{yang2018exposing}, resolution artifacts \cite{li2018exposing}, and absence of eye blinking in videos \cite{li2018ictu} .

% We conduct experiment with standard CNN architectures -- VGG16, ResNet50, ResNet101 and ResNet 152 on self-crafted DeepFake datasets and DARPA challenge. Despite the training data is fully independent to self-crafted DeepFake datasets and DARPA challenge, our network can also achieve satisfied performance.

\begin{figure*}[t]
	\centering
	\includegraphics[width=0.8\linewidth]{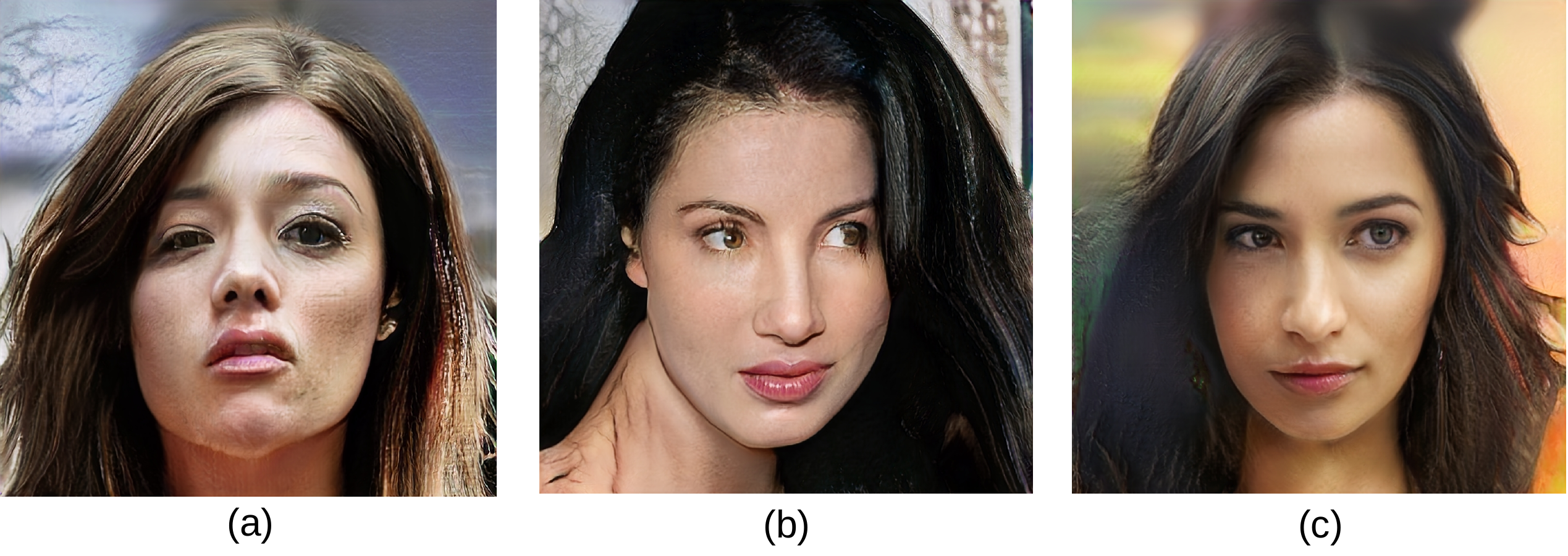}
	~\vspace{-1em}
	\caption{\small \em {Abnormalities of GAN-based faces synthesized by PGGAN \cite{karras2017progressive}. (a) In-symmetric size and location of eyes. (b) Mouth shifts towards left relatively to the nose. (c) Sharp and downwards lateral canthus (inner corner of eye) on left eye.}}

	\label{fig:face_abnormal}
\end{figure*}

%=================================
\section{Related works}
\label{sec:format}
\label{sec:typestyle}

\begin{figure*}[t]
	\centering
	\includegraphics[width=0.9\linewidth]{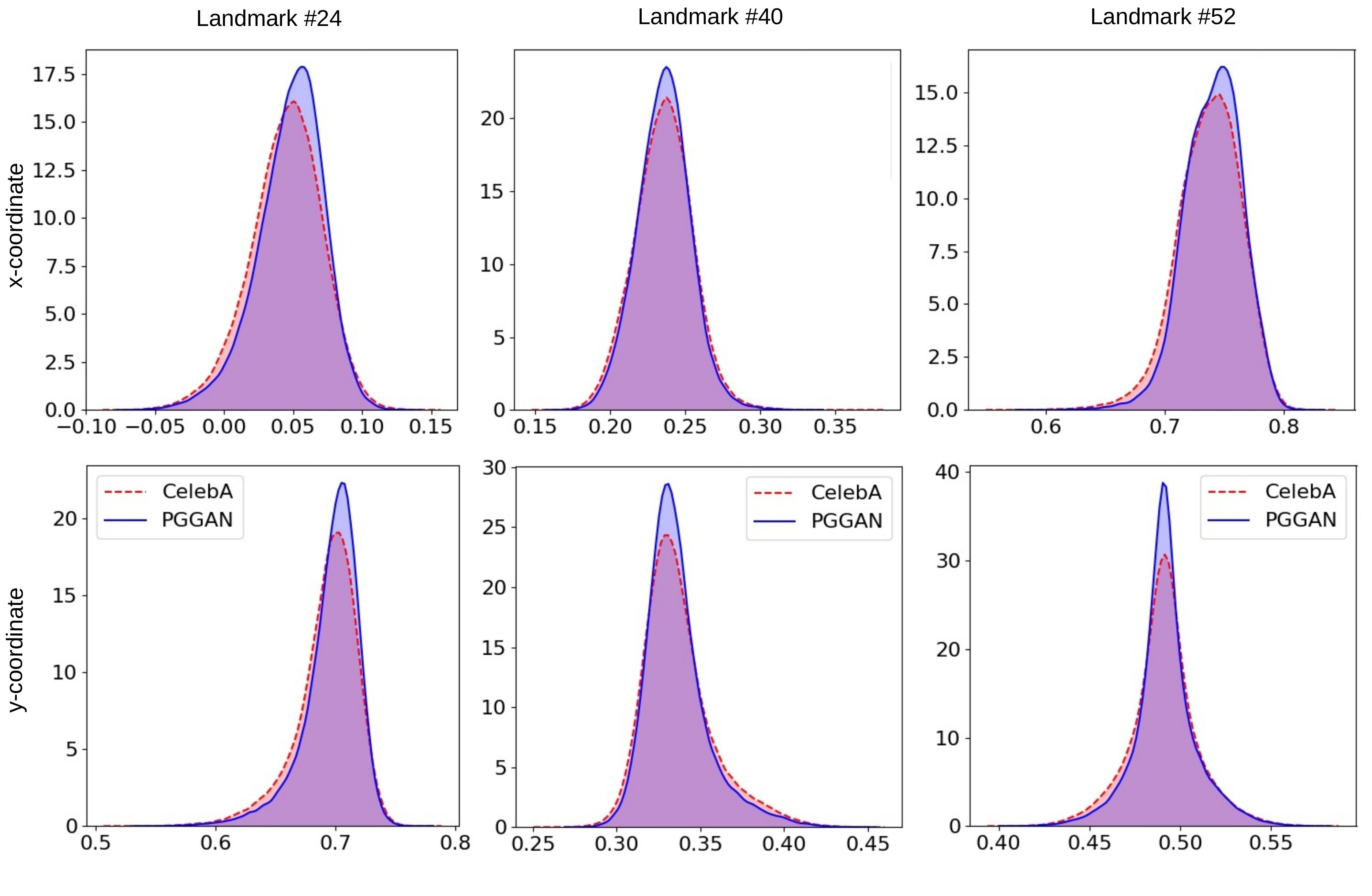}
	~\vspace{-1em}
	\caption{\small \em Density distribution of normalized face landmark locations on real (CelebA) and GAN-synthesized fake (PGGAN) faces over x-, y-coordinates. Real ones are from CelebA dataset with 200k+ images, and fake faces are from PGGAN dataset with 100k images.}
	~\vspace{-2em}
	\label{fig:landmark_dist}
\end{figure*}

\subsection{GAN-based Face Synthesis Methods}

Since the inaugural work of \cite{goodfellow2014generative}, GANs has revolutionized image synthesis methods. A GAN model is consisted of two neural networks, known as the generator and the encoder, that are trained in tandem. The generator takes random noises as input and synthesizes an image, which is sent to a discriminator, aiming to differentiate synthesized images from the real ones. The two networks are trained to compete with each other: the generator aims to create ever more realistic images to defeat the classifier while the discriminator network is trained to be more effective in differentiating the two types of images. The training ends when the two networks reach an equilibrium of the game. The original GAN model has since experienced many improvements. In particular, to improve the 
stability in training, Radford et al. optimized the network architecture by introducing the deep convolutional GANs (DCGAN) \cite{radford2015unsupervised}. Coupled Generative Adversarial Networks (COGAN) learnt joint distribution from different domains further improved realism  of the synthesized images \cite{liu2016coupled}. However, the instability in the training process remains \cite{salimans2016weight, salimans2016improved, berthelot2017began, gulrajani2017improved, kodali2017train}, which propagates to the synthesized samples and limits the model to only synthesize low resolution, see Figure \ref{fig:gan_faces}.

PGGAN \cite{karras2017progressive} is a major breakthrough for synthesizing high resolution realistic face images. It grew both generator and discriminator progressively, starting with generating $4 \times 4$ resolution images from generator. This generated image along with the training image resized into the same scale is feed into the discriminator. After the network are stabilized, a three layer blocks (similar to residual blocks), generating images with doubled heights and widths, faded into the network. These model stabilization through training and higher resolution layers fading in was carried out alternatively, until $1024 \times 1024$ resolution of the generated images is achieved. This approach not only improved training speed and stability, but also synthesized high resolution face images ($1024 \times 1024$) with unprecedented fine details. The PGGAN model is further improved by style-transfer GAN (STGAN) \cite{karras2018style}, which treats face synthesis problem as transferring styles of one face to another. However, STGAN is fundamentally different from previous GAN-based image synthesis models in that images of the best quality are generated conditioned on existing images instead of directly from random noises. Because of this reason, we do not consider detecting STGAN generated images in this work.

\subsection{Detection Methods for GAN-synthesized Images}

Compared to popularity of exploring strategies for synthesizing face images with GANs, methodologies to differentiate the real and synthesized images are far from satisfactory. Li et al \cite{li2018detection} observed the color mismatch in H, S, V and Cb, Cr, Y channels between real and GAN-generated images. Similarly, McCloskey and Albright identified the frequency of saturated pixels and color image statistics of the GAN-generated images are different from the ones captured by cameras \cite{mccloskey2018detecting}. However, this color disparity could easily be removed by post processing after the image synthesis. On the other hand, Mo et al \cite{mo2018fake} and Tariq \cite{tariq2018detecting} designed deep convolutional neural networks classifiers for fake exposure, which usually requires GPU in training and testing, and not be able to reveal the mechanisms behind the classification.

\begin{figure}
	\centering
	\includegraphics[width=0.8\linewidth]{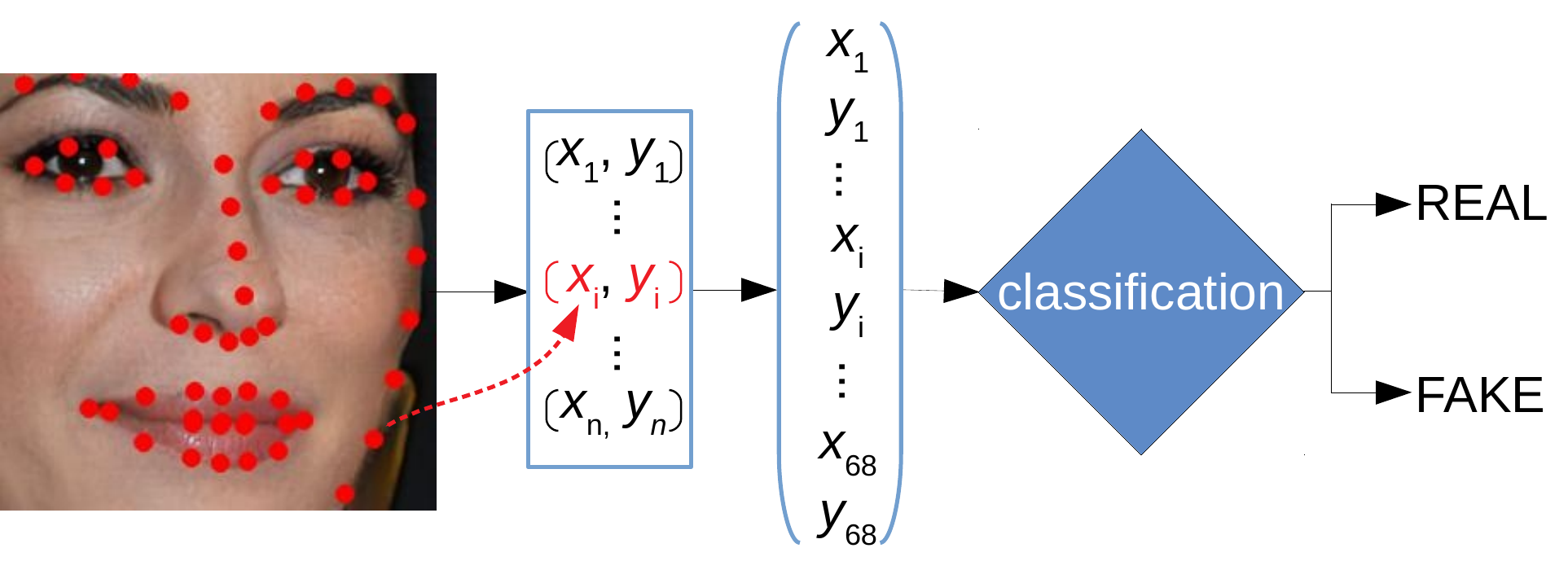}
	~\vspace{-1em}
	\caption{\small \em Pipeline for image our classification method. 68 landmarks are detected from face images which is warped into a standard configuration, followed by flatting landmark locations into 136D vector for classification.}
	\vspace{-0.5cm}
	\label{fig:data_proc}
\end{figure}

\begin{figure*}[t]
	\centering
	\includegraphics[width=0.85\linewidth]{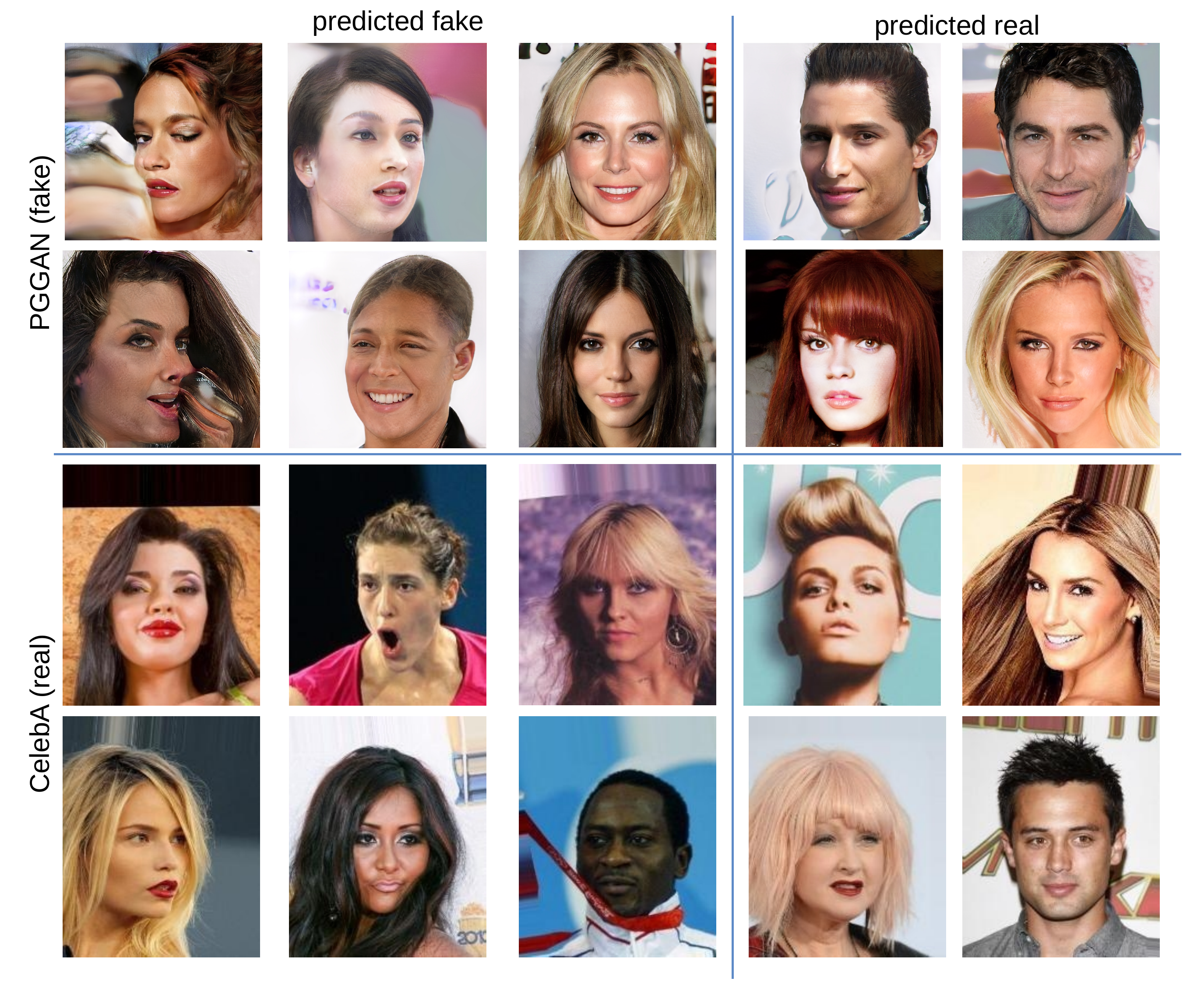}
	~\vspace{-1em}
	\caption{\small \em Examples of correct and incorrect predictions on CelebA and PGGAN datasets.}
	\vspace{0cm}
	\label{fig:pred_example}
\end{figure*}

\section{Method}
\label{sec:method}

As we described in the Introduction, GAN-synthesized faces may exhibit inconsistent configurations of facial parts due to the weak global constraints.  Several examples of this phenomenon are shown in Figure \ref{fig:face_abnormal} for high resolution face images synthesized with the state-of-the-art PGGAN method \cite{karras2017progressive}. In (a), we observe that the synthesized two eyes, nose and upper lips are not symmetric. In (b), the right eye is distorted and the mouth is shifted left-ward with regards to the tip of the nose. In (c), the face shows an unnatural lateral canthus (sharp and down ward inner corner of left eye) and different sizes of two eyes. 

%The observations exemplified in Figure \ref{fig:face_abnormal} suggest that GAN-synthesized faces exhibit abnormalities in the locations of major facial parts due to the lack of global constrains on the facial configuration. 

To quantify such inconsistencies, we compare facial landmark locations detected over GAN-synthesized and real faces. We first run a face detector and extract facial landmarks, Figure \ref{fig:data_proc}. The detected landmarks are warped into a standard configuration in the region of $[0,1] \times [0,1]$ through an affine transformation by minimizing the alignment errors. To reduce the effect of face shape to the alignment result, we follow the standard procedure to estimate the warping transform using only facial landmarks in the central area of the face excluding those on the face contour. Figure \ref{fig:landmark_dist} shows the differences in the aligned landmark locations for the real and GAN-synthesized faces in terms of their distributions along the x- and y- image coordinates. As these results show, the marginal distributions of landmarks for the GAN generated faces exhibit some consistent differences, and such differences are more prominent when we consider the joint distribution over all the coordinates of the ensemble of face landmark points. Therefore, we can use the vector formed by vectorizing all these landmark locations as a feature vector to build a classification system for differentiating GAN-synthesized and real faces, Figure \ref{fig:data_proc}.

There are three advantages of using such features for the classification tasks. First, this feature has relatively low dimension (it is twice the number of landmarks we extract from each face). This facilitates the construction of simpler classification schemes. Second, landmark locations are indifferent of image sizes, so there is no need to rescale the image in training and using the obtained classification method, which may also avoid the undesirable side-effect that leads the classifier to capture the artificial differences in image resolution due to the resizing operation. Third, the abnormalities of facial landmark locations attribute to the underlying fundamental mechanism of GAN image synthesis, so it may not be trivially fixable without introducing more complex constraints into the GAN framework.

\begin{figure}[t]
	\centering
	\includegraphics[width=0.85\linewidth]{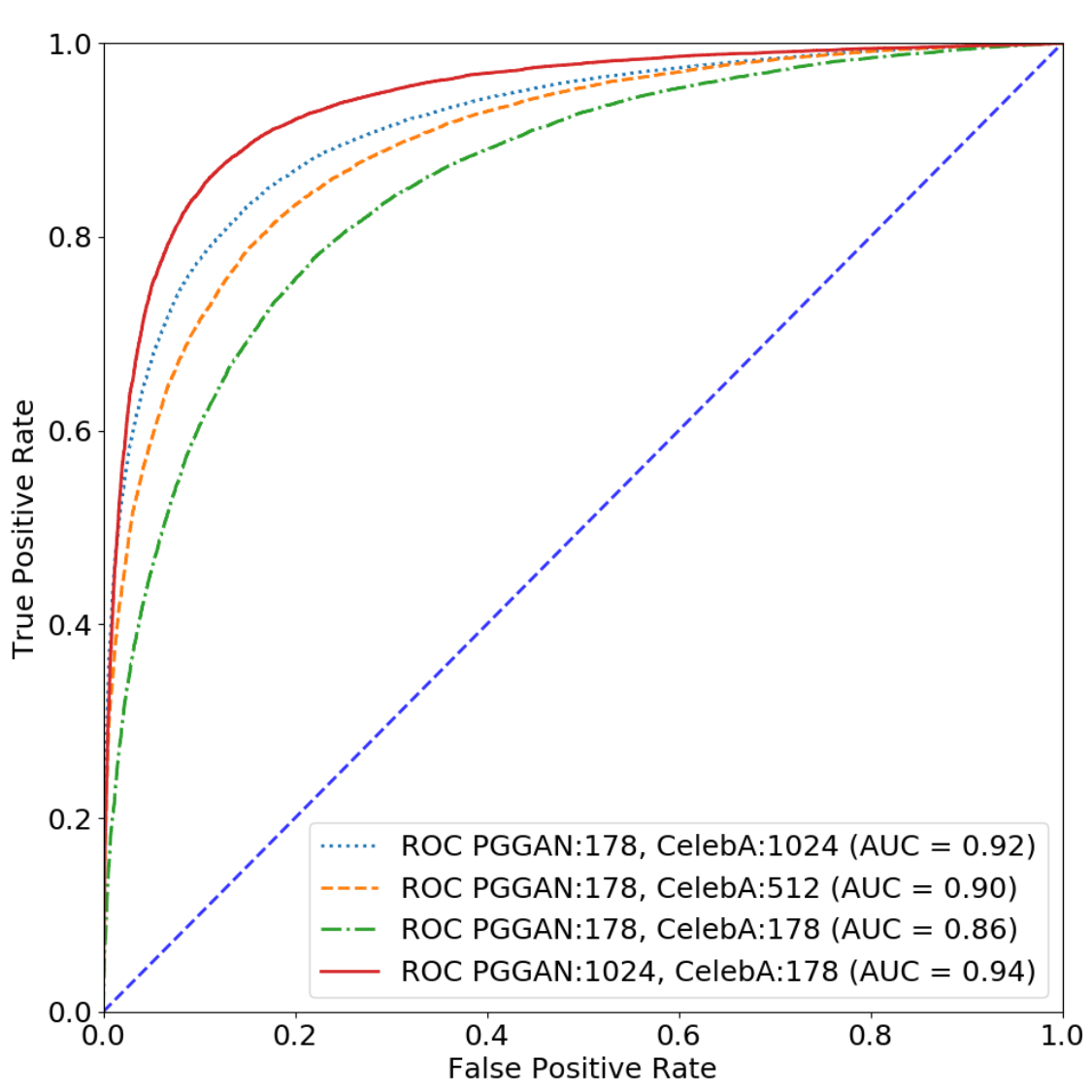}
	~\vspace{-1em}
	\caption{\small Performance by varying the widths of the PGGAN and CelebA images. Widths of resized images for each dataset are in the legend.}
	\vspace{0em}
	\label{fig:varying_width_AUC}
\end{figure}

\begin{table*}[t]
\vspace{1em}
\caption{AUCROC performance of our method and other deep neural network methods \cite{tariq2018detecting} on PGGAN and CelebA resized into different image widths.}
\vspace{-0.3cm}
\begin{tabular}{l|c|c|c|c}
\hline
\multicolumn{1}{c|}{\multirow{2}{*}{method}} & \multirow{2}{*}{\# parameters} & \multicolumn{2}{c|}{dataset (image resolution)}                                                                                                                                              & \multirow{2}{*}{AUROC (\%)} \\ \cline{3-4}
\multicolumn{1}{c|}{}                        &                                & \begin{tabular}[c]{@{}c@{}}CelebA \\ ($216\times178$)\end{tabular}                        & \begin{tabular}[c]{@{}c@{}}PGGAN\\ ($1024\times1024$)\end{tabular}                        &                             \\ \hline
VGG19                                        & $\sim$143.7M                   & \multirow{5}{*}{\begin{tabular}[c]{@{}c@{}}resize to \\ ($1243\times1024$)\end{tabular}} & \multirow{5}{*}{\begin{tabular}[c]{@{}c@{}}remain at\\ ($1024\times1024$)\end{tabular}} & 60.13                       \\ \cline{1-2} \cline{5-5} 
XceptionNet                                  & $\sim$22.9M                    &                                                                                    &                                                                                    & 85.03                       \\ \cline{1-2} \cline{5-5} 
NASNet                                       & $\sim$3.3M                     &                                                                                    &                                                                                    & 96.55                       \\ \cline{1-2} \cline{5-5} 
ShallowNetV3                                 & -                              &                                                                                    &                                                                                    & 99.99                       \\ \cline{1-2} \cline{5-5} 
Our method (SVM)                             & $\sim$110K                     &                                                                                    &                                                                                    & 91.21                       \\ \hline
Our method (SVM)                             & $\sim$110K                     & \begin{tabular}[c]{@{}c@{}}original size \\ ($216\times178$)\end{tabular}                 & \begin{tabular}[c]{@{}c@{}}original size \\ ($1024\times1024$)\end{tabular}               & 94.13                       \\ \hline
\end{tabular}
    \label{table:compareResize}
\end{table*}

\section{Experiments}

% \smallskip
% \begin{table}
% \caption{AUCROC performance by varying the width of the PGGAN images.}
% \vspace{-0.3cm}
%   \begin{tabular}{c|c|c}
%     \hline
%     CelebA width & PGGAN width & AUROC\\ 
%     \hline
%     178 & 178 & 86.10\\
%     \hline
%     178 & 512 & 92.31 \\
%     178 & 1024 & 94.13\\
%     \hline
%     512 & 178 & 89.72 \\
%     1024 & 178 & 91.74\\
%     \hline
%   \end{tabular}
%   \label{table:varyingPGGAN}
%   \vspace{-0.35cm}
% \end{table}

% \begin{figure}[t]
% 	\centering
% 	\includegraphics[width=1\linewidth]{./figure/orfigSizeAUC.pdf}
% 	~\vspace{-1em}
% 	\caption{\small \em ROC curve of SVM classification results on CelebA and PGGAN dataset wtih original figure sizes.}
% 	\vspace{-0.5 cm}
% 	\label{fig:mainResults}
% \end{figure}

In this section, we report the experimental evaluations using landmark locations as features to distinguish real face images from the ones synthesized by GAN.

% compared methods
%GAN and its synthesized images have evolved into several variations since it was firstly introduced. The original GAN framework was used to generate small faces, which is usually blurry and in mono-color \cite{goodfellow2014generative}. Improving the GAN training stability, DCGAN synthesized RGB face images \cite{radford2015unsupervised}, and COGAN incorporated more details on these generated faces \cite{liu2016coupled}. However, visible artifacts can still easily be identified on these face images, such as their unnatrual color, distorted facial components configuration, irregular face profiles and lack of details as in Figure \ref{fig:gan_faces}(a)-(c). Recently, high resolution face images with unprecedented fine details were achieved by Karras et al. \cite{karras2017progressive} as in Figure \ref{fig:gan_faces}(d). The quality of most their images are good enough to deceive our eyes, which make it a challenging dataset for developing image forensics algorithms to expose GAN-synthesized images. A followup work by Karrals et al., STGAN focused on fusing the facial attributes of two existing images  \cite{karras2018style}, hence not in the scope of this study. Therefore, this work  targeted on exposing PGGAN-synthesized images.%Explain here why we only use PGGAN (because early GAN models has low quality and STGAN is not a synthesis algorithm
\smallskip
\noindent {\bf Choosing GAN-based Face Synthesis Method.} Although there are a plenthora of GAN-based face synthesis methods \cite{goodfellow2014generative, radford2015unsupervised, liu2016coupled, karras2017progressive, karras2018style}, we choose in this work the recent PGGAN to construct and evaluate the landmark location based classification method for classifying current state-of-the-art high quality GAN-synthesized faces. This choice is motivated by the following reasons. Early GAN-based face synthesis methods \cite{goodfellow2014generative, radford2015unsupervised, liu2016coupled} produce low quality face images with low resolutions, so they are not representative to the state of the art. On the other hand, the most recent STGAN does not synthesize face images from noise as all other GAN-based method but treat it as a style-transfer problem. 

% Dataset 

\smallskip
\noindent {\bf Dataset.} The training and testing of the SVM classifiers for exposing GAN-synthesized images are based on two datasets: (a) CelebFaces Attributes Dataset (CelebA) \cite{liu2015faceattributes} contains more than 200K real face images with fixed resolution of $216 \times 178$ pixels. (b) PGGAN dataset \cite{karras2017progressive} consisting of 100K PGGAN-synthesized face images at a resolution of $1024 \times 1024$ pixels are used as fake faces. $75\%$ of both datasets are merged as negative and positive samples for training, and the rest $25\%$ are used for testing.

\smallskip
\noindent {\bf Preprocessing and Training.} Using the normalized locations of all face landmarks as features, we can develop a simple classification scheme to differentiate the real and GAN-synthesized faces, with standard classification methods such as SVM or neural networks, Figure \ref{fig:data_proc}. In this study, the normalized landmark locations of each face ($\in R^{68\times2}$) are flattened in to a vector ($\in R^{136\times1}$), which is standardized by subtracting the mean and divided by the standard deviation of all training samples. We trained SVM classifiers with radial basis function (RBF) kernel with a grid search on the hyperparameters using 5-fold cross validation. The losses of two classes are balanced by adjusting the sample loss inversely proportional to class frequencies in training dataset. 

% qualitative results
\smallskip
\noindent{\bf Performance.} Figure \ref{fig:pred_example} shows some examples of prediction results on PGGAN and CelebA datasets. PGGAN-synthesized faces with artifacts could be correctly predicted as fake faces, and the ones falsely predicted to be real mostly bears no visible defects. For the real face images in the CelebA dataset, some faces are falsely predicted to be fake. This may result from difficulties in accurately estimating landmark locations in faces with strong facial expression and occlusion, as shown in the bottom left figures in Figure \ref{fig:pred_example}. 

% quantitative results

Quantititave results of our method is shown in Table \ref{table:compareResize}, in terms of the Area Under ROC (AUROC). As a comparison, we also include performance with different neural network architectures from \cite{tariq2018detecting} on the same dataset\footnote{These results are taken from the published paper \cite{tariq2018detecting} directly, because no code is currently available. The training and testing data may differ.}. Note that all methods in \cite{tariq2018detecting} take the image as input. To accommodate the different sizes of input images between the CelebA  ($216 \times 178$) and PGGAN ($1024\times 1024$) datasets, the images are resized to the same size and the results on enlarging the celebA images to $1243 \times 1024$ are reported in Table \ref{table:compareResize}. 

As the results show, the SVM classifier achieves an AUROC of 94.13\% and outperforms several deep neural network based methods (\eg, VGG19 and XceptionNet). The two deep neural network based methods achieving higher classification accuracy are with much higher number of parameters. More importantly, these results are obtained on resized images and no study was conducted on the effect of the resizing on the final classification -- as upsampling an image lead to certain artifacts. It is not clear how much of the high performance can be attribute to the intrinsic difference between the two types of images. As we mentioned previously, the feature based on locations of facial landmarks is independent from image sizes and we compare the effect of resizing the two classes of images in another set of experiments shown in Figure \ref{fig:varying_width_AUC}, which demonstrates that the classification performance is relatively indifferent to the resizing operation. We would also like to emphasize that all these CNN models requires GPU for training and testing, while our method has much fewer parameters and only CPU for training and testing.

\begin{figure*}[t]
	\centering
	\includegraphics[width=0.8\linewidth]{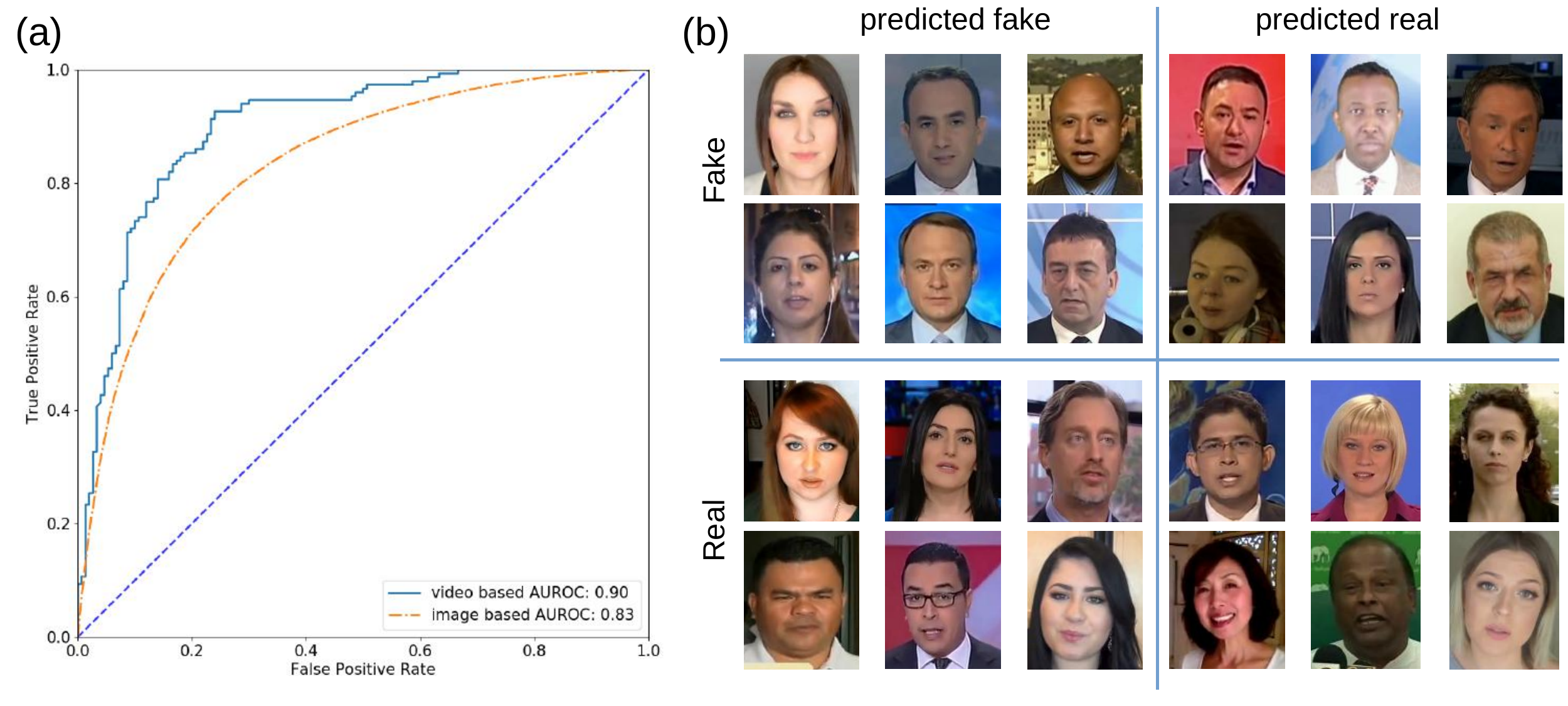}
	~\vspace{-1em}
	\caption{\small \em Classification on FaceForensics Testing Dataset. (a) The ROC curve and AUROC scores. (b) Examples of correct and incorrect predictions for both classes.}
	\vspace{0cm}
	\label{fig:FF_pred_example}
\end{figure*}

\section{Results on Face Forensics}

Although the feature we proposed is originally designed for GAN-based face synthesis, we believe that other types of face synthesis methods may also exhibit similar abnormalities. To this end, we test our method on the FaceForensics dataset, which contains pairs of real and falsified videos synthesized by Face2Face \cite{thies2016face2face}. It has 740 pairs of videos (726,270 images) for training, 150 pairs (151,052 images) for validation, and 150 videos (155,490 images) for testing \cite{rossler2018faceforensics}. The video frames in FaceForensics dataset vary in $576$ to $1920$ pixels for width and $480$ to $1080$ pixels for height. SVM models based on the landmark features are trained similarly on this dataset and we report the performance in Figure \ref{fig:FF_pred_example}, which achieves a 0.83 AUROC for classification in individual frames. By averaging the classification prediction on individual videos, the AUROC increases to 0.90.

% \smallskip
% \begin{table}
% \caption{AUCROC performance by fixing width of videos in different scales for FaceForensics Dataset.}
% \vspace{-0.3cm}
%   \begin{tabular}{c|c|c|c}
%     \hline
%     fake width & real width & image AUROC & video AUROC\\ 
%     \hline
%     256 & 256 & 0.75 & 0.80\\
%     512 & 512 & 0.82 & 0.90 \\
%     1024 & 1024 & 0.81 & 0.87\\
%     \hline
%   \end{tabular}
%   \label{table:varyingPGGAN}
%   \vspace{-0.35cm}
% \end{table}

% \begin{figure}[t]
% 	\centering
% 	\includegraphics[width=0.6\linewidth]{./figure/landmark_on_faces.pdf}
% 	~\vspace{-1em}
% 	\caption{\small \em 68 facial landmarks.}
% 	~\vspace{-2em}
% 	\label{fig:landmarks}
% \end{figure}

% \subsection{Ablation Study} 
% performance on compression

\section{Conclusion}
In this work, we proposed using aligned facial landmark locations as features to distinguish PGGAN synthesized fake human face images. Our method is based on the observation that current GAN-based algorithms uses random noises as input, which is good at depicting the details of face parts, but lack of constrains on the configuration of different face components. Consequently, it introduces errors in facial parts locations, which is non-trivial to be fixed in GAN models. We performed experiments to demonstrate this phenomenon and further developed classification models on this cue. The results indicated the effectiveness of our methods with low dimensional input, light-weighted models, and robust to scale variation. 

\bibliographystyle{ACM-Reference-Format}
\bibliography{refs}


\begin{thebibliography}{21}

%%% the Web address.

\ifx \showCODEN    \undefined \def \showCODEN     #1{\unskip}     \fi
\ifx \showDOI      \undefined \def \showDOI       #1{#1}\fi
\ifx \showISBNx    \undefined \def \showISBNx     #1{\unskip}     \fi
\ifx \showISBNxiii \undefined \def \showISBNxiii  #1{\unskip}     \fi
\ifx \showISSN     \undefined \def \showISSN      #1{\unskip}     \fi
\ifx \showLCCN     \undefined \def \showLCCN      #1{\unskip}     \fi
\ifx \shownote     \undefined \def \shownote      #1{#1}          \fi
\ifx \showarticletitle \undefined \def \showarticletitle #1{#1}   \fi
\ifx \showURL      \undefined \def \showURL       {\relax}        \fi
% invisible to TeX
\providecommand\bibfield[2]{#2}
\providecommand\bibinfo[2]{#2}
\providecommand\natexlab[1]{#1}
\providecommand\showeprint[2][]{arXiv:#2}

\bibitem[\protect\citeauthoryear{Arulkumaran, Deisenroth, Brundage, and
  Bharath}{Arulkumaran et~al\mbox{.}}{2017}]%
        {arulkumaran2017brief}
\bibfield{author}{\bibinfo{person}{Kai Arulkumaran},
  \bibinfo{person}{Marc~Peter Deisenroth}, \bibinfo{person}{Miles Brundage},
  {and} \bibinfo{person}{Anil~Anthony Bharath}.}
  \bibinfo{year}{2017}\natexlab{}.
\newblock \showarticletitle{A brief survey of deep reinforcement learning}.
\newblock \bibinfo{journal}{\emph{arXiv preprint arXiv:1708.05866}}
  (\bibinfo{year}{2017}).
\newblock


\bibitem[\protect\citeauthoryear{Berthelot, Schumm, and Metz}{Berthelot
  et~al\mbox{.}}{2017}]%
        {berthelot2017began}
\bibfield{author}{\bibinfo{person}{David Berthelot}, \bibinfo{person}{Thomas
  Schumm}, {and} \bibinfo{person}{Luke Metz}.} \bibinfo{year}{2017}\natexlab{}.
\newblock \showarticletitle{Began: Boundary equilibrium generative adversarial
  networks}.
\newblock \bibinfo{journal}{\emph{arXiv preprint arXiv:1703.10717}}
  (\bibinfo{year}{2017}).
\newblock


\bibitem[\protect\citeauthoryear{Deng}{Deng}{2014}]%
        {deng2014tutorial}
\bibfield{author}{\bibinfo{person}{Li Deng}.} \bibinfo{year}{2014}\natexlab{}.
\newblock \showarticletitle{A tutorial survey of architectures, algorithms, and
  applications for deep learning}.
\newblock \bibinfo{journal}{\emph{APSIPA Transactions on Signal and Information
  Processing}}  \bibinfo{volume}{3} (\bibinfo{year}{2014}).
\newblock


\bibitem[\protect\citeauthoryear{Goodfellow, Bengio, and Courville}{Goodfellow
  et~al\mbox{.}}{2016}]%
        {Goodfellow-et-al-2016}
\bibfield{author}{\bibinfo{person}{Ian Goodfellow}, \bibinfo{person}{Yoshua
  Bengio}, {and} \bibinfo{person}{Aaron Courville}.}
  \bibinfo{year}{2016}\natexlab{}.
\newblock \bibinfo{booktitle}{\emph{Deep Learning}}.
\newblock \bibinfo{publisher}{MIT Press}.
\newblock
\newblock
\shownote{\url{http://www.deeplearningbook.org}.}


\bibitem[\protect\citeauthoryear{Goodfellow, Pouget-Abadie, Mirza, Xu,
  Warde-Farley, Ozair, Courville, and Bengio}{Goodfellow et~al\mbox{.}}{2014}]%
        {goodfellow2014generative}
\bibfield{author}{\bibinfo{person}{Ian Goodfellow}, \bibinfo{person}{Jean
  Pouget-Abadie}, \bibinfo{person}{Mehdi Mirza}, \bibinfo{person}{Bing Xu},
  \bibinfo{person}{David Warde-Farley}, \bibinfo{person}{Sherjil Ozair},
  \bibinfo{person}{Aaron Courville}, {and} \bibinfo{person}{Yoshua Bengio}.}
  \bibinfo{year}{2014}\natexlab{}.
\newblock \showarticletitle{Generative adversarial nets}. In
  \bibinfo{booktitle}{\emph{NIPS}}.
\newblock


\bibitem[\protect\citeauthoryear{Gulrajani, Ahmed, Arjovsky, Dumoulin, and
  Courville}{Gulrajani et~al\mbox{.}}{2017}]%
        {gulrajani2017improved}
\bibfield{author}{\bibinfo{person}{Ishaan Gulrajani}, \bibinfo{person}{Faruk
  Ahmed}, \bibinfo{person}{Martin Arjovsky}, \bibinfo{person}{Vincent
  Dumoulin}, {and} \bibinfo{person}{Aaron~C Courville}.}
  \bibinfo{year}{2017}\natexlab{}.
\newblock \showarticletitle{Improved training of wasserstein gans}. In
  \bibinfo{booktitle}{\emph{Advances in Neural Information Processing
  Systems}}. \bibinfo{pages}{5767--5777}.
\newblock


\bibitem[\protect\citeauthoryear{Karras, Aila, Laine, and Lehtinen}{Karras
  et~al\mbox{.}}{2017}]%
        {karras2017progressive}
\bibfield{author}{\bibinfo{person}{Tero Karras}, \bibinfo{person}{Timo Aila},
  \bibinfo{person}{Samuli Laine}, {and} \bibinfo{person}{Jaakko Lehtinen}.}
  \bibinfo{year}{2017}\natexlab{}.
\newblock \showarticletitle{Progressive growing of gans for improved quality,
  stability, and variation}.
\newblock \bibinfo{journal}{\emph{arXiv preprint arXiv:1710.10196}}
  (\bibinfo{year}{2017}).
\newblock


\bibitem[\protect\citeauthoryear{Karras, Laine, and Aila}{Karras
  et~al\mbox{.}}{2018}]%
        {karras2018style}
\bibfield{author}{\bibinfo{person}{Tero Karras}, \bibinfo{person}{Samuli
  Laine}, {and} \bibinfo{person}{Timo Aila}.} \bibinfo{year}{2018}\natexlab{}.
\newblock \showarticletitle{A style-based generator architecture for generative
  adversarial networks}.
\newblock \bibinfo{journal}{\emph{arXiv preprint arXiv:1812.04948}}
  (\bibinfo{year}{2018}).
\newblock


\bibitem[\protect\citeauthoryear{Kodali, Abernethy, Hays, and Kira}{Kodali
  et~al\mbox{.}}{2017}]%
        {kodali2017train}
\bibfield{author}{\bibinfo{person}{Naveen Kodali}, \bibinfo{person}{Jacob
  Abernethy}, \bibinfo{person}{James Hays}, {and} \bibinfo{person}{Zsolt
  Kira}.} \bibinfo{year}{2017}\natexlab{}.
\newblock \showarticletitle{How to train your DRAGAN}.
\newblock \bibinfo{journal}{\emph{arXiv preprint arXiv:1705.07215}}
  \bibinfo{volume}{2}, \bibinfo{number}{4} (\bibinfo{year}{2017}).
\newblock


\bibitem[\protect\citeauthoryear{Li, Li, Tan, and Huang}{Li
  et~al\mbox{.}}{2018}]%
        {li2018detection}
\bibfield{author}{\bibinfo{person}{Haodong Li}, \bibinfo{person}{Bin Li},
  \bibinfo{person}{Shunquan Tan}, {and} \bibinfo{person}{Jiwu Huang}.}
  \bibinfo{year}{2018}\natexlab{}.
\newblock \showarticletitle{Detection of deep network generated images using
  disparities in color components}.
\newblock \bibinfo{journal}{\emph{arXiv preprint arXiv:1808.07276}}
  (\bibinfo{year}{2018}).
\newblock


\bibitem[\protect\citeauthoryear{Liu and Tuzel}{Liu and Tuzel}{2016}]%
        {liu2016coupled}
\bibfield{author}{\bibinfo{person}{Ming-Yu Liu} {and} \bibinfo{person}{Oncel
  Tuzel}.} \bibinfo{year}{2016}\natexlab{}.
\newblock \showarticletitle{Coupled generative adversarial networks}. In
  \bibinfo{booktitle}{\emph{Advances in neural information processing
  systems}}. \bibinfo{pages}{469--477}.
\newblock


\bibitem[\protect\citeauthoryear{Liu, Luo, Wang, and Tang}{Liu
  et~al\mbox{.}}{2015}]%
        {liu2015faceattributes}
\bibfield{author}{\bibinfo{person}{Ziwei Liu}, \bibinfo{person}{Ping Luo},
  \bibinfo{person}{Xiaogang Wang}, {and} \bibinfo{person}{Xiaoou Tang}.}
  \bibinfo{year}{2015}\natexlab{}.
\newblock \showarticletitle{Deep Learning Face Attributes in the Wild}. In
  \bibinfo{booktitle}{\emph{Proceedings of International Conference on Computer
  Vision (ICCV)}}.
\newblock


\bibitem[\protect\citeauthoryear{McCloskey and Albright}{McCloskey and
  Albright}{2018}]%
        {mccloskey2018detecting}
\bibfield{author}{\bibinfo{person}{Scott McCloskey} {and}
  \bibinfo{person}{Michael Albright}.} \bibinfo{year}{2018}\natexlab{}.
\newblock \showarticletitle{Detecting GAN-generated Imagery using Color Cues}.
\newblock \bibinfo{journal}{\emph{arXiv preprint arXiv:1812.08247}}
  (\bibinfo{year}{2018}).
\newblock


\bibitem[\protect\citeauthoryear{Mo, Chen, and Luo}{Mo et~al\mbox{.}}{2018}]%
        {mo2018fake}
\bibfield{author}{\bibinfo{person}{Huaxiao Mo}, \bibinfo{person}{Bolin Chen},
  {and} \bibinfo{person}{Weiqi Luo}.} \bibinfo{year}{2018}\natexlab{}.
\newblock \showarticletitle{Fake Faces Identification via Convolutional Neural
  Network}. In \bibinfo{booktitle}{\emph{Proceedings of the 6th ACM Workshop on
  Information Hiding and Multimedia Security}}. ACM, \bibinfo{pages}{43--47}.
\newblock


\bibitem[\protect\citeauthoryear{Radford, Metz, and Chintala}{Radford
  et~al\mbox{.}}{2015}]%
        {radford2015unsupervised}
\bibfield{author}{\bibinfo{person}{Alec Radford}, \bibinfo{person}{Luke Metz},
  {and} \bibinfo{person}{Soumith Chintala}.} \bibinfo{year}{2015}\natexlab{}.
\newblock \showarticletitle{Unsupervised representation learning with deep
  convolutional generative adversarial networks}.
\newblock \bibinfo{journal}{\emph{arXiv preprint arXiv:1511.06434}}
  (\bibinfo{year}{2015}).
\newblock


\bibitem[\protect\citeauthoryear{R{\"o}ssler, Cozzolino, Verdoliva, Riess,
  Thies, and Nie{\ss}ner}{R{\"o}ssler et~al\mbox{.}}{2018}]%
        {rossler2018faceforensics}
\bibfield{author}{\bibinfo{person}{Andreas R{\"o}ssler},
  \bibinfo{person}{Davide Cozzolino}, \bibinfo{person}{Luisa Verdoliva},
  \bibinfo{person}{Christian Riess}, \bibinfo{person}{Justus Thies}, {and}
  \bibinfo{person}{Matthias Nie{\ss}ner}.} \bibinfo{year}{2018}\natexlab{}.
\newblock \showarticletitle{Faceforensics: A large-scale video dataset for
  forgery detection in human faces}.
\newblock \bibinfo{journal}{\emph{arXiv preprint arXiv:1803.09179}}
  (\bibinfo{year}{2018}).
\newblock


\bibitem[\protect\citeauthoryear{Salimans, Goodfellow, Zaremba, Cheung,
  Radford, and Chen}{Salimans et~al\mbox{.}}{2016}]%
        {salimans2016improved}
\bibfield{author}{\bibinfo{person}{Tim Salimans}, \bibinfo{person}{Ian
  Goodfellow}, \bibinfo{person}{Wojciech Zaremba}, \bibinfo{person}{Vicki
  Cheung}, \bibinfo{person}{Alec Radford}, {and} \bibinfo{person}{Xi Chen}.}
  \bibinfo{year}{2016}\natexlab{}.
\newblock \showarticletitle{Improved techniques for training gans}. In
  \bibinfo{booktitle}{\emph{Advances in neural information processing
  systems}}. \bibinfo{pages}{2234--2242}.
\newblock


\bibitem[\protect\citeauthoryear{Salimans and Kingma}{Salimans and
  Kingma}{2016}]%
        {salimans2016weight}
\bibfield{author}{\bibinfo{person}{Tim Salimans} {and} \bibinfo{person}{Durk~P
  Kingma}.} \bibinfo{year}{2016}\natexlab{}.
\newblock \showarticletitle{Weight normalization: A simple reparameterization
  to accelerate training of deep neural networks}. In
  \bibinfo{booktitle}{\emph{Advances in Neural Information Processing
  Systems}}. \bibinfo{pages}{901--909}.
\newblock


\bibitem[\protect\citeauthoryear{Tariq, Lee, Kim, Shin, and Woo}{Tariq
  et~al\mbox{.}}{2018}]%
        {tariq2018detecting}
\bibfield{author}{\bibinfo{person}{Shahroz Tariq}, \bibinfo{person}{Sangyup
  Lee}, \bibinfo{person}{Hoyoung Kim}, \bibinfo{person}{Youjin Shin}, {and}
  \bibinfo{person}{Simon~S Woo}.} \bibinfo{year}{2018}\natexlab{}.
\newblock \showarticletitle{Detecting both machine and human created fake face
  images in the wild}. In \bibinfo{booktitle}{\emph{Proceedings of the 2nd
  International Workshop on Multimedia Privacy and Security}}. ACM,
  \bibinfo{pages}{81--87}.
\newblock


\bibitem[\protect\citeauthoryear{Thies, Zollhofer, Stamminger, Theobalt, and
  Nie{\ss}ner}{Thies et~al\mbox{.}}{2016}]%
        {thies2016face2face}
\bibfield{author}{\bibinfo{person}{Justus Thies}, \bibinfo{person}{Michael
  Zollhofer}, \bibinfo{person}{Marc Stamminger}, \bibinfo{person}{Christian
  Theobalt}, {and} \bibinfo{person}{Matthias Nie{\ss}ner}.}
  \bibinfo{year}{2016}\natexlab{}.
\newblock \showarticletitle{Face2face: Real-time face capture and reenactment
  of rgb videos}. In \bibinfo{booktitle}{\emph{Proceedings of the IEEE
  Conference on Computer Vision and Pattern Recognition}}.
  \bibinfo{pages}{2387--2395}.
\newblock


\bibitem[\protect\citeauthoryear{Zhang, Yang, Chen, and Li}{Zhang
  et~al\mbox{.}}{2018}]%
        {zhang2018survey}
\bibfield{author}{\bibinfo{person}{Qingchen Zhang}, \bibinfo{person}{Laurence~T
  Yang}, \bibinfo{person}{Zhikui Chen}, {and} \bibinfo{person}{Peng Li}.}
  \bibinfo{year}{2018}\natexlab{}.
\newblock \showarticletitle{A survey on deep learning for big data}.
\newblock \bibinfo{journal}{\emph{Information Fusion}}  \bibinfo{volume}{42}
  (\bibinfo{year}{2018}), \bibinfo{pages}{146--157}.
\newblock


\end{thebibliography}
\end{document}